\documentclass[utf8]{FrontiersinHarvard} % for articles in journals using the Harvard Referencing Style (Author-Date), for Frontiers Reference Styles by Journal: https://zendesk.frontiersin.org/hc/en-us/articles/360017860337-Frontiers-Reference-Styles-by-Journal
\usepackage{url,hyperref,lineno,microtype,subcaption,amsmath}
\usepackage[onehalfspacing]{setspace}

\def\keyFont{\fontsize{8}{11}\helveticabold }
\def\firstAuthorLast{Mészáros {et~al.}} %use et al only if is more than 1 author
\def\Authors{Balázs Mészáros\,$^{1,*}$, James Knight\,$^{1}$ and Thomas Nowotny\,$^{1}$}
\def\Address{$^{1}$Sussex AI, School of Engineering and Informatics, University of Sussex, Brighton, United Kingdom}
\def\corrAuthor{Balázs Mészáros}

\def\corrEmail{United Kingdom, Falmer, Brighton BN1 9RH, bm452@sussex.ac.uk}

\begin{document}
\onecolumn
\firstpage{1}

\title[Learning Delays Through Gradients and Structure]{Learning Delays Through Gradients and Structure: Emergence of Spatiotemporal Patterns in Spiking Neural Networks} 

%Title1: Dynamic Pruning and Delay Learning in Spiking Neural Networks: Spatio-temporal Patterns of Excitation and Inhibition
%Title2: Learning Delays Through Gradients and Through Structure: Emergence of Spatiotemporal Patterns in Spiking Neural Networks
%Title3: Spatio-temporal Pattern Emergence in Spiking Neural Networks: Do Delays Need Gradients?
%Old title: Spatio-temporal Structure of Excitation and Inhibition Emerges in Spiking Neural Networks with and without Biologically Plausible Constraints

\author[\firstAuthorLast ]{\Authors} %This field will be automatically populated
\address{} %This field will be automatically populated
\correspondance{} %This field will be automatically populated

\extraAuth{}% If there are more than 1 corresponding author, comment this line and uncomment the next one.
%\extraAuth{corresponding Author2 \\ Laboratory X2, Institute X2, Department X2, Organization X2, Street X2, City X2 , State XX2 (only USA, Canada and Australia), Zip Code2, X2 Country X2, email2@uni2.edu}

\maketitle

\begin{abstract}

%%% Leave the Abstract empty if your article does not require one, please see the Summary Table for full details.
\section{}
We present a Spiking Neural Network (SNN) model that incorporates learnable synaptic delays through two approaches: per-synapse delay learning via Dilated Convolutions with Learnable Spacings (DCLS) and a dynamic pruning strategy that also serves as a form of delay learning. In the latter approach, the network dynamically selects and prunes connections, optimizing the delays in sparse connectivity settings. We evaluate both approaches on the Raw Heidelberg Digits keyword spotting benchmark using Backpropagation Through Time with surrogate gradients.

Our analysis of the spatio-temporal structure of synaptic interactions reveals that, after training, excitation and inhibition group together in space and time. Notably, the dynamic pruning approach, which employs DEEP R for connection removal and RigL for reconnection, not only preserves these spatio-temporal patterns but outperforms per-synapse delay learning in sparse networks. 

Our results demonstrate the potential of combining delay learning with dynamic pruning to develop efficient SNN models for temporal data processing. Moreover, the preservation of spatio-temporal dynamics throughout pruning and rewiring highlights the robustness of these features, providing a solid foundation for future neuromorphic computing applications.

\tiny
 \keyFont{ \section{Keywords:} Spiking Neural Network, Delay Learning, Dynamic Pruning, Receptive Field, Sparse Connectivity} %All article types: you may provide up to 8 keywords; at least 5 are mandatory.
\end{abstract}

\section{Introduction}

Spiking Neural Networks (SNNs) are the third generation of artificial neural networks~\citep{maass1997networks}, inspired by the functioning of biological neurons. Unlike traditional neural networks, which are \emph{stateless} and process information through continuous activation values, neurons in SNNs are \emph{stateful} and communicate via sparse binary spikes, mimicking the electrical impulses observed in biological neurons. This enables SNNs to efficiently process temporal information, making them well-suited to tasks involving sequential data processing. SNNs have shown significant potential as a computational paradigm for neuromorphic computing platforms~\citep{furber2014spinnaker, davies2018loihi, merolla2014million}, enabling low-power and real-time processing. This makes them a compelling basis for next-generation intelligent systems.

Synaptic delays (between the emission of a spike and its arrival at the post-synaptic neuron) have been suggested as a means of improving the spatio-temporal information processing of SNNs~\citep{izhikevich2006polychronization,paugam2008delay}. These delays, which can vary across connections, allow neurons to perform coincidence detections across longer time intervals, enhancing SNN's ability to process temporal information. In biological systems, delays encompass axonal, synaptic, and dendritic components and are modified by processes like myelination to facilitate learning~\citep{bengtsson2005extensive} and coincidence detection~\citep{seidl2010mechanisms}. Furthermore,  neuromorphic hardware such as Intel's Loihi~\citep{davies2018loihi}, SpiNNaker~\citep{furber2014spinnaker} and DenRAM\citep{d2024denram} incorporate programmable synaptic delays, enabling SNNs with delays to be efficiently deployed for real-time data processing. In the past, two kinds of delay learning methods have been used: delay selection and delay shift. Delay selection relies on implementing several synapses between each neuron with various delays, and picking the most optimal one~\citep{bohte2002error}. Delay shift uses only a single synapse, and optimizes the corresponding delay. An early example of delay shift was the Delay Learning Remote Supervised Method (DL-ReSuMe) that showed improved performance compared to just training weights both in terms of accuracy and training speed~\citep{taherkhani2015dl}. \citet{wang2019delay} further improved upon these results and \citet{shrestha2018slayer} extended the approach to enable synaptic delay learning in deep networks. 
In this paper, we use Dilated Convolutions with Learnable Spacings (DCLS)~\citep{khalfaoui2021dilated} which, similarly to the approach proposed by \citeauthor{wang2019delay}, convolves spike trains with delay kernels. However, DCLS can be used in deeper architectures and has been shown to be an effective delay learning method on neuromorphic benchmarks~\citep{hammouamri2023learning}. 

Methods to decrease the number of parameters in a neural network were first proposed more than 35 years ago~\citep{lecun1989optimal, hassibi1993optimal} and, in recent years, overparametrisation in deep neural networks has become a widely acknowledged problem~\citep{ba2014deep}.  The lottery ticket
hypothesis proposes that, within an overparameterised dense network, there exist multiple sparse sub-networks with varying performances and, among them, one sub-network stands out as the ``winning ticket'' that outperforms the others~\citep{frankle2018lottery}. Sparse neural networks significantly reduce memory usage and energy consumption and are  \emph{required} on many neuromorphic systems, which often have a maximum fan-in per neuron~\citep{schemmel2010wafer,merolla2014million} or limited memory available for connectivity~\citep{furber2014spinnaker,davies2018loihi}. 
One means of obtaining a sparse network is to train a dense network and remove connections using a process called ``pruning''~\citep{han2015learning}. However, this means that the size of models is still limited by the training cost of the original dense model. In contrast, biological systems dynamically rewire synaptic connections during learning, suggesting that dynamic pruning (also known as structure learning) and rewiring can enhance neural network performance and efficiency. The first algorithm that both disconnected and reconnected neurons during training was DEEP~R~\citep{bellec2018deep}. The method drops synapses based on their weight changes during learning and replaces them with randomly chosen synapses to maintain a constant number of synapses. In parallel, another dynamic pruning method called sparse evolutionary training (SET) was introduced, relying purely on weight magnitudes to drop connections~\citep{mocanu2018scalable}. Sparse Networks from Scratch~(SNFS) used the momentum of each parameter as the criterion for reconnecting
neurons~\citep{dettmers2019sparse}. RigL~\citep{evci2021rigging} took this one step further and uses gradient information for growing the network. These dynamic pruning methods aim to emulate biological efficiency, potentially offering superior sparsity and accuracy with fewer floating-point operations (FLOPs). According to calculations by~\citeauthor{evci2021rigging}, at 90\% sparsity, RigL requires 1/4 of FLOPs compared to the same size dense model, and DEEP R requires 1/10. With implementations exploiting sparsity, both algorithms can significantly speed up training and inference~\citep{knight2023easy}. 

In this paper, we present our analysis of a spiking neural network trained in a supervised fashion on the Heidelberg Digits benchmark dataset~\citep{Cramer_2022}, preprocessed as described by \citet{zenke2021remarkable}. We analyze the learnt parameters of the fully connected network, then train networks with dynamic pruning and fixed sparse connectivity, and conduct the same analysis on these more efficient and biologically more plausible architectures.
\section{Methods}
We consider an SNN with Leaky-Integrate-and-Fire (LIF) neurons, solved with a linear Euler method,
\begin{eqnarray}
    u_i^{(l)}[t] &=& \left(1-\frac{\Delta t}{\tau}\right)u_i^{(l)}[t-1]\left(1-S_i^{(l)}[t-1]\right) +I_i^{(l)}[t] \cdot \Delta t \label{eq:euler} \\
    S_j^{(l)}[t] &=& \Theta(u_j^{(l)}[t]-\vartheta),\label{eq:spike}
\end{eqnarray}
where, $u_i^{(l)}[t]$ is the membrane potential of the $i$-th neuron in layer $l$ at time step $t$, $\tau=10.05$ is the membrane time constant and $I_i^{(l)}(t)$ is the input current. $S_i^{(l)}$ is the spike train emitted by neuron $i$, $\vartheta$ denotes the firing threshold and $\Theta$ the Heaviside function. 
For our numerical experiments, we used a timestep of $\Delta t = 1$. 

Because $\Theta$ is non-differentiable, we replace it with a $\arctan$ surrogate gradient during the backward pass of our training~\citep{neftci2019surrogate}. 
To implement the synaptic delay training, the input current $I_i^{(l)}(t)$ is calculated using DCLS~\citep{hammouamri2023learning}, i.e. convolving the spike train $S_j^{(l-1)}$ from layer $l-1$ with the 1D kernel
%\begin{equation}
%    I_i^{(l)}=\sum_j k_{ij}^{(l)}*S_j^{(l-1)}\label{eq:kernel_input}
%\end{equation}
\begin{equation}
    k_{ij}^{(l)}[n] = \frac{w_{ij}^{(l)}}{c}\exp\left(-\frac{1}{2}\left(\frac{n-T_d+d_{ij}^{(l)}+1}{\sigma}\right)^2\right), \quad n = 0, \ldots, T_d
    \label{eq:kernel}
\end{equation}
where $T_d=25$ is the maximum delay, $d_{ij}^{(l)}$ is the synaptic delay from neuron $j$ to neuron $i$ in layer $l$, $c$ is a normalization term so that $\sum_{n = 0}^{T_d} k_{ij}^{(l)}[n] = w_{ij}^{(l)}$ and $\sigma=12.5$ is the standard deviation of the delay kernel, which is decreased during training. As the kernel $k_{ij}^{(l)}$ slides through the spike train $S_{j}^{(l-1)}$ at each timestep $t$ the kernel will have access to spikes in the range of \{$t-T_d,\ldots , t$\}. The method essentially creates a temporal convolutional kernel, where the position of weights in the kernel corresponds to the synaptic delay. These weights can then be learned as normal, providing a framework in which weights and delays can be optimised together. For more details, we refer the reader to the original publication~\citep{hammouamri2023learning}.

For dynamic pruning, we combined two methods: DEEP R~\citep{bellec2018deep} for dropping connections and RigL~\citep{evci2021rigging} for introducing them.  In DEEP R the synaptic weights are defined as $w_{ij}^{l} = s_{ij}^{(l)} \text{max}(\theta_{ij}^{(l)},0)$, where $s_{ij}^{(l)}$ is a constant sign value, and $\theta_{ij}^{(l)}$ is the parameter trained with gradient descent. If $\theta_{ij}^{(l)}$ is not positive, the connection is considered dormant. We use $L1$ regularization to encourage pruning of unnecessary connections. Although the original DEEP R method adds noise to the gradients to induce stochasticity in ANNs, following~\citet{bellec2020solution}, we omitted this in our experiments. DEEP R maintains a fixed number of synapses by randomly reactivating synapses throughout training.
\begin{figure}
    \begin{center}
    \includegraphics[width=180mm]{rf_full.png}
    \end{center}
    \caption{\label{fig:rf}(A) Toy example of how the spatio-temporal receptive fields were generated. $W$ denotes the synaptic strength, $D$ denotes the synaptic delay. (B) The receptive field with the highest observed Moran's I value prior to training (left) and after training (middle) in a dense network. The distributions (right) show all observed Moran's I values. (C) The receptive field with the highest observed Moran's I value prior to training (left) and after training (middle) in a sparse network. The distributions (right) show all observed Moran's I values. }
\end{figure}
RigL introduces synapses based on gradient analysis rather than randomly. In each iteration, the algorithm selects the $k$ strongest negative gradients from the inactive connections:
\begin{equation}
    \operatornamewithlimits{ArgTopK}_{\theta^{(l)}_{ij} \leq 0}\left(-\frac{d L}{d\theta^{(l)}_{ij}}, k\right).
\end{equation}
Since we make the assumption that the weight of active connections is positive, the original RigL method needs to be modified slightly. Instead of picking synapses to introduce based on the \emph{absolute}  gradient value, we simply pick based on the strongest negative gradient values, since a positive gradient implies that gradient descent wants to keep the connection inactive.

Most neural network models do not adhere to Dale's law, which states that neurons are either exclusively excitatory or inhibitory. Computationally, this means that the signs for the outgoing weights from each presynaptic neuron are the same. Since with DEEP R we have to generate the sign matrix before training, we can conveniently apply this constraint by generating a random sign vector $s\in\{{-1,+1}\}$ and broadcasting it into a matrix. Unless stated otherwise, all of our results apply to networks that adhere to Dale's law.

To measure the spatial autocorrelation of the learned spatio-temporal patterns, we used the method of Moran's I~\citep{moran1950notes}:
\begin{equation}
    I=\frac{N}{W}\frac{\sum_{i=1}^{N}\sum_{j=1}^{N}w_{ij}(x_i-\bar{x})(x_j-\bar{x})}{\sum_{i=1}^{N}(x_i-\bar{x})^2}\label{eq:moransi}
\end{equation}
where $N$ is the number of elements, $x$ are the elements in the pattern, $\bar{x}$ is the mean of the elements, $w_{ij}$ are the elements of the spatial weights with zero diagonals, and $W$ is the sum of all $w_{ij}$. We used the $8$ neighbourhood case (also known as the Queen's case) for the weights $w_{ij}$ to capture a broad influence. Since the ordering in the spatial dimension is arbitrary (i.e. the ordering of the rows (neurons) can be changed), we took 2000 random row permutations of the matrix and determined the maximum Moran's I across them. With no spatial autocorrelation Moran's I is $\frac{-1}{N-1}$, meaning that it approaches $0$ with increasing $N$. Moran's $I$ values significantly below $\frac{-1}{N-1}$ indicates negative spatial autocorrelation whereas those significantly above $\frac{-1}{N-1}$ indicate positive spatial autocorrelation. We measure if the distributions of Moran's~I values in trained and untrained networks are significantly different using the Mann-Whitney U test.

We extended the PyTorch implementation of DCLS developed by \citet{hammouamri2023learning} for our experiments. Our architecture consisted of one hidden layer with 256 neurons, dropout layers with a probability of 0.4, batch normalization, and delays in all layers in the range of $({0, \ldots, 25})$ timesteps. We used a voltage sum readout and cross-entropy loss for training and our model was trained with the Adam optimizer, using a OneCycle scheduler for weights and a Cosine Annealing scheduler for delays.
\section{Results}
\subsection{Spatio-temporal receptive fields of trained networks}
In our experiments, we trained our models with the best hyperparameters used by~\citet{hammouamri2023learning} for training on the Spiking Heidelberg Digits (SHD) dataset, but instead trained on the Raw Heidelberg Digits dataset~\citep{Cramer_2022}. We preprocessed the dataset similarly to \citet{zenke2021remarkable}, creating Mel spectrograms of shape 40 (input channels) by 80 (timesteps). We performed our experiments on this dataset because it carries complex enough temporal information to benefit from delay learning, but it can be solved with a relatively small network, which makes it better suited for our analysis. The only additional parameter we needed to tune was the $L1$ regularization strength. We first trained our model without any sparsity or Dale's law-derived constraints and achieved $97.2$\%. This is higher than the results reported by \citet{zenke2021remarkable} ($94\pm2$\%) -- the only other paper running benchmarks on RawHD. However, the goal of this paper is not benchmarking. We then analysed the learnt spatio-temporal patterns by extracting what we refer to as ``spatio-temporal receptive fields'' using the process illustrated in Figure \ref{fig:rf}A. While, in most settings, receptive fields refer to \emph{functional} attributes of the network \citep{linden2003spectrotemporal, deangelis1995receptive}, we use a similar concept to analyse the \emph{structural} attributes of a trained network in terms of the sign and delay of connections. For each hidden neuron, we created a panel with input neurons on the y-axis and delay along the x-axis. Then, for each input neuron, we place one point at the x coordinate corresponding to the learnt delay  of its connection to the hidden neuron and colored either blue or red based on the weight’s sign. We then summed these panels for each output layer neuron, weighting each by the learned hidden-to-output weight and aligning them on the x-axis according to the learned hidden-to-output delay. These figures allow us to see whether a certain feature for a given class is excited or inhibited and whether it takes effect immediately or with a delay. The results for the dense network before and after training are shown in Figure \ref{fig:rf}B. Visually, the figure illustrates that spatio-temporal patterns of excitation and inhibition formed after training.
\begin{figure}
    \begin{center}
    \includegraphics[width=95mm]{strucutre_vs_delay_example.png}
    \end{center}
    \caption{\label{fig:structure_vs_delay_toy} Types of learning. Delays are indicated by the thickness of the connections, and red is used to highlight the changes.}
\end{figure}
\subsection{Spatio-temporal autocorrelation of learned receptive fields}
We assessed the spatio-temporal autocorrelation of each of the 20 output neuron's trained receptive fields by calculating the maximum Moran’s I from 2000 random row-wise permutations (as the spatial ordering of neurons in our architecture is arbitrary). We repeated our experiment 3 times and created distributions from the $3\times20$ receptive fields for trained and untrained networks (see Figure \ref{fig:rf}C). To assess whether there is a meaningful difference in spatio-temporal correlation within receptive fields between trained and untrained networks, we analyzed the distributions of Moran’s I values for both network types using a Mann-Whitney U test. This non-parametric test was chosen because it does not assume a specific distribution shape for the data, making it suitable for comparing independent samples with potentially different variances and non-normality. For the dense networks, the Mann-Whitney U test yielded a test statistic of $3598.0$ and a p-value of approximately $3.88\times10^{-21}$. Given this very low p-value, we reject the null hypothesis (H0) that the distributions of Moran’s I values in trained and untrained dense networks are identical. This result indicates a highly statistically significant difference between the two distributions with the trained dense networks exhibiting systematically different spatial correlations than the untrained networks. 
\begin{figure}
    \begin{center}
    \includegraphics[width=180mm]{new_structure_vs_delay.jpg}
    \end{center}
    \caption{\label{fig:structure_vs_delay}Comparing the effects of structure learning and delay learning. Learning the structure does not make a huge difference with lower sparsity, but as it increases the benefit becomes clear. This might not be that surprising, since these methods were built to train highly sparse models. Delay learning improves the results with a fixed connectivity matrix, as was also shown in~\citep{hammouamri2023learning}. The benefits of delay learning are not obvious when the structure is learnt.}
\end{figure}
\subsection{Dynamic pruning}
Next, we trained networks with dynamic pruning -- utilizing DEEP R and RigL to enforce a fixed level of 87.5\% sparsity and Dale's law by making all of each neuron's outgoing connections have the same sign. While, visually, the emergence of grouping is not as obvious as it was in the dense networks (Figure \ref{fig:rf}C), calculating Moran's I and performing a similar Mann-Whitney U test produced a test statistic of $3169.0$ and a p-value of approximately $6.69\times10^{-13}$. Although this p-value is higher than that observed for the dense network, it still provides strong evidence to reject the null hypothesis, indicating a statistically significant difference between the trained and untrained distributions for the sparse network.

These findings suggest that training introduces structural features within the network’s receptive fields, contributing to increased spatial correlation that is absent in untrained networks. While the effect is more pronounced in dense networks, the presence of a significant difference in the sparse network, despite its high level of sparsity, highlights that spatial correlations remain an important outcome of the training process. This may imply that the network’s learning captures and reinforces spatial dependencies critical to its task, as reflected in the elevated Moran’s I values in trained networks.

\subsection{Ablation study}
In this section, we analyse the combined and separate usefulness of dynamic pruning (i.e. structure learning) and delay learning. 
We defined four models, one where the structure is learnt, one where the synaptic delays are learnt, one where both are learnt and one where neither are (the synaptic weights are still trained in all cases). See Figure \ref{fig:structure_vs_delay_toy} for illustration.
For the simulations with fixed connectivity, we omitted Dale's law, since for the models with fixed connectivity, poor initialisation could impose significant constraints on the network.
Starting with a 50\% sparsity, we progressively halved the number of connections in a sequence of experiments. The effectiveness of DEEP R and RigL seems highly dependent on initialization, so we ran three repeats of each configuration and reported the average classification accuracy. Figure \ref{fig:structure_vs_delay} demonstrates that dynamic pruning is highly effective, especially as sparsity increases. While learning delays is beneficial when the structure is fixed, its benefits are less obvious when the structure is learnt.
Overall, dynamic pruning seems vital for maintaining high performance in sparse networks and we would argue that, when structure is learned, delay learning might not be necessary.

\section{Discussion}
%In this work, we have studied the effects of learning synaptic delays and sparse connectivity in SNNs. With synaptic delays present, neuron's receptive fields can be visualised spatio-temporally, analogous to the way we examine the purely spatial receptive fields of CNNs. These learnt delays enable us to understand what the network captures in the spatio-temporal domain. We showed visually that excitatory and inhibitory groupings emerge with various network setups. However, Moran's I only confirmed this in the fully connected architecture.

%\citet{hammouamri2023learning} previously demonstrated the benefits of learning delays when the connectivity matrix is sparse. Here, we delved deeper into this topic by combining learnable delays with dynamic pruning. We combined DEEP R -- which in the non-stochastic setting has only one hyperparameter and makes it convenient to incorporate Dale's law -- with RigL, which utilises the gradient to decide which synapses should be introduced into the network.

%We also investigated how dense and sparse networks compare if we fix the number of trainable parameters. We found that in a setting where the number of synapses is the same for a sparse network and a dense network, the sparse network performs better even when Dale’s principle is enforced.

%Finally, we investigated the combined and separate effects of delay and structure learning. 
SNNs have a strong potential for spatio-temporal processing and synaptic delays only enhance this. Our study demonstrates a new approach for the visual analysis of networks with delay learning, revealing that functional spatio-temporal patterns emerge in both dense and sparse networks. These patterns appear more in networks where the framework allows for the optimization of temporal parameters through gradients or structural learning. We compared the Moran's I distributions of models trained with and without delay learning and dynamic pruning. The mean Moran's I value of the model with fixed delays and structure was much lower  (0.027 compared to 0.064) and the distributions were significantly different (Mann-Whitney U test statistic of $166.5$ and p-value of approximately $9.97\times10^{-18}$). At the same time the classification performance is less (see Figure \ref{fig:structure_vs_delay}, red line), suggesting that emerging spatio-temporal structure and classification success correlate.

While delay learning seems useful with sparse connectivity, we found that learning the structure is more important. In fact, when the structure was learnt, delay learning appeared to bring little benefit. This may be because gradient descent struggles with simultaneously learning both parameters; if a newly connected synapse has a non-optimized delay, it might be immediately deactivated again. However, our experiments with using RigL for removing connections (which does not happen every epoch like DEEP R) did not show significant benefits, challenging this theory. From another perspective, structure learning with fixed delays \emph{is} a form of delay learning since, if a synapse has an ineffective delay, the network can adapt by introducing a more effective synapse with a new random delay.
This method closely resembles ``delay select'' SNNs~\citep{bohte2002error}. However, for this method to work in the fully connected setting, several connections are required between each pre and postsynaptic neuron, with the number increasing as the delay distribution widens. Ergo, from the point of view of efficiency, we have an argument for delay learning methods such as the one based on DCLS, especially in the fully connected setting. However, we argue that a good test for the usefulness of a given delay learning method is whether, in a sparse network, it performs better than simply replacing synapses with new ones that have a fixed random delay. In our experiments, we relied on gradient information to reconnect weights, but this only yielded minor improvements over random reconnection (standard DEEP R). This implies that, in our simulations, randomly sampling a delay value was just as effective as adjusting delays using gradient descent. While both in our experiments and the ones run by~\citet{hammouamri2023learning}, delay learning does improve network performance, perhaps the precision that delay gradient information provides is not necessary and slows down learning performance. 

In conclusion, our results show that sparse network architectures can be efficient for machine learning tasks. Our findings pave the way for future research into the optimization of SNNs for various applications, particularly those at the edge that have strong memory and computational constraints.

\section*{Conflict of Interest Statement}
%All financial, commercial or other relationships that might be perceived by the academic community as representing a potential conflict of interest must be disclosed. If no such relationship exists, authors will be asked to confirm the following statement: 

The authors declare that the research was conducted in the absence of any commercial or financial relationships that could be construed as a potential conflict of interest.

\section*{Author Contributions}

%The Author Contributions section is mandatory for all articles, including articles by sole authors. If an appropriate statement is not provided on submission, a standard one will be inserted during the production process. The Author Contributions statement must describe the contributions of individual authors referred to by their initials and, in doing so, all authors agree to be accountable for the content of the work. Please see  

%\href{https://www.frontiersin.org/about/policies-and-publication-ethics#AuthorshipAuthorResponsibilities}{here} for full authorship criteria.
BM -- Conceptualization, Investigation, Formal Analysis, Software, Validation, Visualization, Writing - original draft, Writing - review \& editing \\
JCK -- Conceptualization, Supervision, Funding acquisition, Writing - review \& editing \\
TN -- Conceptualization, Supervision, Funding acquisition, Writing - review \& editing

\section*{Funding}
This work was supported through a Doctoral Scholarship by the Leverhulme Trust and the EPSRC (EP/S030964/1, EP/V052241/1).

%\section*{Acknowledgments}
%This is a short text to acknowledge the contributions of specific colleagues, institutions, or %agencies that aided the efforts of the authors.

%\section*{Supplemental Data}
% \href{http://home.frontiersin.org/about/author-guidelines#SupplementaryMaterial}{Supplementary Material} should be uploaded separately on submission, if there are Supplementary Figures, please include the caption in the same file as the figure. LaTeX Supplementary Material templates can be found in the Frontiers LaTeX folder.

\section*{Data Availability Statement}
Our source code is available on \href{https://github.com/mbalazs98/spatiotemporal}{GitHub}.
% Please see the availability of data guidelines for more information, at https://www.frontiersin.org/about/author-guidelines#AvailabilityofData

\bibliographystyle{Frontiers-Harvard} %  Many Frontiers journals use the Harvard referencing system (Author-date), to find the style and resources for the journal you are submitting to: https://zendesk.frontiersin.org/hc/en-us/articles/360017860337-Frontiers-Reference-Styles-by-Journal. For Humanities and Social Sciences articles please include page numbers in the in-text citations 
\bibliography{test}

%%% Make sure to upload the bib file along with the tex file and PDF
%%% Please see the test.bib file for some examples of references

%\section*{Figure captions}

%%% Please be aware that for original research articles we only permit a combined number of 15 figures and tables, one figure with multiple subfigures will count as only one figure.
%%% Use this if adding the figures directly in the mansucript, if so, please remember to also upload the files when submitting your article
%%% There is no need for adding the file termination, as long as you indicate where the file is saved. In the examples below the files (logo1.eps and logos.eps) are in the Frontiers LaTeX folder
%%% If using *.tif files convert them to .jpg or .png
%%%  NB logo1.eps is required in the path in order to correctly compile front page header %%%

%%% If you don't add the figures in the LaTeX files, please upload them when submitting the article.
%%% Frontiers will add the figures at the end of the provisional pdf automatically
%%% The use of LaTeX coding to draw Diagrams/Figures/Structures should be avoided. They should be external callouts including graphics.

\end{document}